\pdfoutput=1

\documentclass[11pt]{article}

\usepackage[]{acl}

\usepackage{times}
\usepackage{latexsym}

\usepackage[T1]{fontenc}

\usepackage[utf8]{inputenc}

\usepackage{microtype}

\usepackage{graphicx}
\usepackage{mathtools}
\usepackage{multirow}
\usepackage{bm}
\usepackage{hyperref}       %
\usepackage{url}            %
\usepackage{booktabs}       %
\usepackage{amsfonts}       %
\usepackage{nicefrac}       %
\DeclareMathOperator*{\argmax}{argmax}

\newcommand{\zsteer}{{\boldsymbol{z}_{steer}}}
\usepackage[bottom]{footmisc}
\usepackage{titlesec}
\usepackage{xurl}
\usepackage[linesnumbered,ruled]{algorithm2e}
\usepackage{booktabs}
\usepackage{makecell}
\usepackage[normalem]{ulem}

\newcommand{\xvar}[1]{\textsf{#1}}

\newcommand{\xvbox}[2]{\makebox[#1][l]{#2}}

\SetAlFnt{\footnotesize}

\SetAlCapSty{xAlCapSty}

\SetCommentSty{xCommentSty}

\SetNlSty{mynlfont}{}{} 

\LinesNumbered

\SetSideCommentRight

\DontPrintSemicolon

\RestyleAlgo{algoruled}

\title{Extracting Latent Steering Vectors from Pretrained Language Models}

\renewcommand{\textsuperscript}[1]{\raisebox{0.5ex}{#1}}

\author{Nishant Subramani\textsuperscript{$\dagger$} \quad Nivedita Suresh\textsuperscript{$\diamondsuit$} \quad Matthew E.\ Peters\textsuperscript{$\dagger$} \\ \\
\textsuperscript{$\dagger$}Allen Institute for Artificial Intelligence, Seattle, WA, USA\\
\textsuperscript{$\diamondsuit$}Arrive Bio, San Francisco, CA, USA \\
\texttt{\{nishants,matthewp\}@allenai.org} \\
{\texttt{\{nive\}@arrivebio.com}}
}

\begin{document}
\maketitle

\begin{abstract}
Prior work on controllable text generation has focused on \emph{learning} how to control language models through trainable decoding, smart-prompt design, or fine-tuning based on a desired objective. 
We hypothesize that the information needed to steer the model to generate a target sentence is already encoded within the model.
Accordingly, we explore a different approach altogether: \emph{extracting} latent vectors directly from pretrained language model decoders without fine-tuning.
Experiments show that there exist \textit{steering vectors}, which, when added to the hidden states of the language model, generate a target sentence nearly perfectly (> 99 BLEU) for English sentences from a variety of domains.
We show that vector arithmetic can be used for unsupervised sentiment transfer on the Yelp sentiment benchmark, with performance comparable to models tailored to this task.
We find that distances between steering vectors reflect sentence similarity when evaluated on a textual similarity benchmark (STS-B), outperforming pooled hidden states of models.
Finally, we present an analysis of the intrinsic properties of the steering vectors.
Taken together, our results suggest that frozen LMs can be effectively controlled through their latent steering space.~\footnote{Code is available at \url{https://github.com/nishantsubramani/steering_vectors}.}

\end{abstract}

\begin{figure*}[t]\centering
\includegraphics[width=\textwidth]{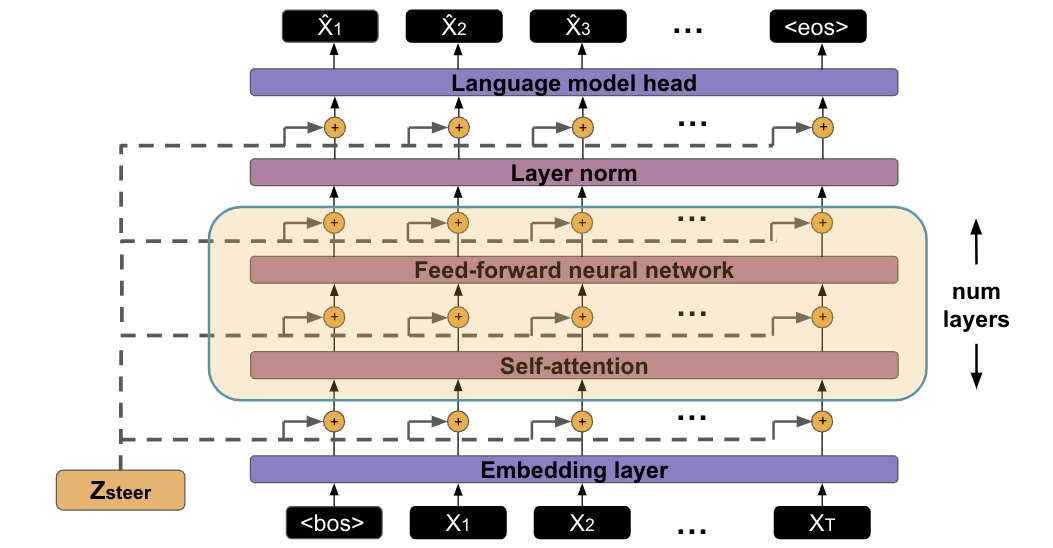}
\caption{Our approach adds a vector $\zsteer$ to the activations of a pretrained transformer decoder to steer it to decode a desired target sentence. We experiment with adding $\zsteer$ to different locations inside a GPT-2 model at different timesteps. Experiments reveal that our approach can recover sequences nearly perfectly and that injecting the steering vector in the middle layers of the transformer stack performs best. Layer normalizations and residual connections inside the transformer block are omitted for clarity. 
}
\label{fig:process}
\end{figure*}

\section{Introduction}
Leveraging large pretrained language models trained on massive Web corpora has become the go-to approach to solve natural language processing tasks~\cite{Peters2018DeepCW, radford2018improving, Devlin2018BERTPO, Brown2020LanguageMA}.
As a result, controlling these models has become paramount as many applications of NLP technology require control over the generations of the model.
Prior work aims to \textit{learn} how to control language models and falls in three categories: trainable decoding~\cite{gu-etal-2017-trainable, Deng2020ResidualEM}, smart-prompt design~\cite{shin-etal-2020-autoprompt, lester-etal-2021-power}, and fine-tuning based on a desired objective~\cite{krause-etal-2021-gedi-generative, weng2021conditional}.
Further, many works opt to train auto-encoder based models for controllable text generation~\cite{Shen2017StyleTF, Shen2020EducatingTA, mai-etal-2020-plug}.
These approaches make controllability easier by learning a latent space that is more easily manipulated to encourage models to generate text corresponding to a target attribute such as positive sentiment in the case of sentiment transfer.

We take a more direct approach and explore whether it is possible to \emph{extract} latent vectors directly from pretrained language model decoders without fine-tuning.
We call these vectors \textit{steering vectors} and define the \textit{latent steering space} of a sentence under a language model by the set of extracted steering vectors, which steer the model to generate that sentence exactly.
During decoding, we add our steering vector to the hidden states of the language model to generate the target sentence.
Rather than training a model to learn steering vectors, we provide several methods to \emph{extract} fixed-length steering vectors directly from pretrained language model decoders.
Experiments show that we can extract steering vectors effectively, achieving nearly perfect recovery for English sentences from a variety of domains without fine-tuning the underlying language model at all.

Next, we take our extracted steering vectors and explore whether they can be used for unsupervised sentiment transfer on the Yelp sentiment benchmark~\citep{zhang2015character}.
We find that adding an offset vector to extracted steering vectors performs comparably to carefully designed, autoencoder-based models.
To see whether steering vectors encode semantics, we explore whether they can be used for unsupervised textual similarity. %
On the semantic textual similarity benchmark (STS-B,~\citet{cer-etal-2017-semeval}), our steering vectors outperform extractive methods such as averaging language model hidden states and GloVe vectors~\citep{pennington-etal-2014-glove} when measuring the cosine similarity between vectors, but fall short of lexical methods tailored to semantic similarity tasks and methods that finetune on natural language inference datasets. %

Lastly, we analyze the intrinsic properties of the latent space of our steering vectors.
Experiments show that decoding from interpolations in the latent space produces meaningful output, and that steering vectors from different domains cluster together.
Also, we find that our methods do not simply memorize the target sequence like a naive compression algorithm, and instead leverage the model.
Taken together, our results suggest that frozen language models can be controlled effectively through their latent steering space.

\section{Extracting Steering Vectors}
This section discusses our method for extracting a steering vector for a target sentence from a frozen, pretrained language model.
Throughout this paper, we use GPT2 as our language model and use its 117M parameter model size~\cite{Radford2019LanguageMA}, although our approach can be directly applied to any transformer-based autoregressive language model decoder~\cite{vaswani2017attention}.

\subsection{Steering Vectors}
In controllable text generation and textual style transfer, prior work based on denoising and variational autoencoders opt for a disentangling approach.
These approaches encode the source sequence into a fixed-length vector using an encoder, apply style transformations using a controller, and finally decode from the transformed vector using a decoder~\cite{Shen2017StyleTF, Jin2020DeepLF}.
Instead of learning an encoder and controller to uncover a representation, we ask whether its possible to extract a vector directly from a pretrained language model decoder in order to steer the model.

Due to the success of hidden layer manipulations for language models including adapter-based fine-tuning~\cite{Houlsby2019ParameterEfficientTL}, plug-and-play language models~\cite{dathathri2019plug}, and offset-vector-based recovery and style transfer among others~\cite{subramani2019can, Shen2020EducatingTA, mai-etal-2020-plug, montero-etal-2021-sentence}, we choose to manipulate the hidden states as well.

Our method works by adding a fixed-length vector $\zsteer$ to the hidden states of a pretrained and frozen LM.  For a desired target sentence, we randomly initialize $\zsteer$ and optimize it via gradient descent to maximize the likelihood of the model given the target sentence. At decoding time, we feed a $\zsteer$ to the model and perform decoding as usual.
The choice of a fixed-length vector makes analysis more meaningful, allowing us to compare vectors for different sentences with different lengths in the same representation space.

\subsection{Discovering steering vectors}
We define our steering vectors $\zsteer \in \mathbb{R}^{d'}$.
In our experiments, $d' \leq d$, where $d$ is the hidden dimension of the underlying language model (for GPT2-117M, $d=768$).
If $d' < d$, we project $\zsteer$ using a semi-orthogonal matrix, $W_{steer} \in \mathbb{R}^{d' \times d}$, which preserves scale. $W_{steer}$ is initialized randomly, never trained, and never updated.

We estimate a steering vector $\hat{z}_{steer} \in \mathbb{R}^{d'}$ via the language model for a sentence $\boldsymbol{x}$ by maximizing the log probability of $x$, while keeping the language model fixed:
\begin{equation}
\label{eq:sentence-projection}
    \hat{\boldsymbol{z}}_{steer} = \argmax_{\zsteer \in \mathcal{Z}} \sum_{t=1}^T \log p(x_t | \boldsymbol{x}_{<t}, \zsteer)
\end{equation}
Here, $\mathcal{Z} \in \mathbb{R}^{d'}$.
Note: we find a single steering vector $\zsteer$ for each sentence $\boldsymbol{x}$. %
We use stochastic gradient descent with the Adam~\cite{kingma2014adam} optimizer and cross entropy loss to find the best $\hat{\boldsymbol{z}}_{steer}$, while freezing the language model.
See algorithm~\ref{alg: extract_steering_vector} for the pseudocode.

Since our method adds $\zsteer$ to the activations of the model, the layer we add $\zsteer$ to affects recoverability. We experiment with injecting $\zsteer$ at different layers (\textit{injection locations)}: at the embedding layer, right before language model head (LM Head), after self-attention layer in the transformer stack, after feed-forward layer in the transformer stack as well as combinations of them.
In addition to varying injection locations, we also vary the timesteps where $\zsteer$ gets added. 
We experiment with adding $\zsteer$ at just the first timestep and at every timestep.
See Figure~\ref{fig:process} for details.

\begin{algorithm}
\SetKwFunction{cumprod}{cumprod}
\SetKwFunction{length}{length}
\SetKwFunction{zeros}{zeros}
\SetKwFunction{ceil}{ceil}

\SetKwInOut{Input}{Input}
\SetKwInOut{Output}{Output}

\caption{Extracting $\zsteer$ for a sentence\label{alg: extract_steering_vector}}

\Input{%
		\xvbox{2.5mm}{$\xvar{x}$} -- target sentence \\
		\xvbox{2.5mm}{$\xvar{M}$} -- pretrained language model \\
		\xvbox{2.5mm}{$\theta$} -- pretrained language model weights \\
		\xvbox{2.5mm}{${I_L}$} -- injection location \\
		\xvbox{2.5mm}{${I_T}$} -- injection timestep \\
		\xvbox{2.5mm}{$\xvar{d}$} -- dimension of $\zsteer$ \\
	  }
\Output{%
		\xvbox{8mm}{$\zsteer$} -- extracted candidate steering vector 
	   }

  \BlankLine %
  
  \xvbox{2mm}{$\zsteer \sim$ xavier\_normal(d)}\\
  \For{$\xvar{i} \leftarrow \xvar{[1, 2, ..., N]}$} {
    \xvbox{1mm}{$\xvar{logits} = \xvar{M}_{\theta}.forward(\xvar{x}, \zsteer, {I_L}, {I_T})$}
    
    \xvbox{1mm}{$\mathcal{L} = XENT(logits, x)$}
    
    \xvbox{1mm}{$\mathcal{L}.backward()$}
    
    \xvbox{1mm}{$\zsteer = \zsteer + lr * \frac{\partial{\mathcal{L}}}{\partial{\zsteer}}$}
  }
  \xvbox{2mm}{return $\zsteer$}
\end{algorithm}

\subsection{Steering Language Models}
We steer the language model using $\zsteer$ to generate a target sentence $\boldsymbol{x}$ by passing in a beginning-of-sentence token and $\zsteer$ to the model.
Since we are interested in exact generation, all results presented use greedy decoding without assuming a true length.
We stop when decoding produces an end-of-sentence token or produces 1024 tokens, the maximum length that GPT-2 can generate. 

\section{Can we extract steering vectors?}
Here, we show that we can robustly extract steering vectors that generate target sentences perfectly.

\subsection{Experimental setup}
We gather a broad corpus spanning four different domains and measure the extent to which our approach can extract a steering vector for each sentence under a variety of experimental conditions, where we vary injection locations and timesteps. %

\paragraph{Data Collection}
For these experiments on sentence recoverability, we create a dataset which combines four corpora from different domains: movie dialogs (movies), classic books (books), news articles (news), and Wikipedia (wiki).
For movies, we choose the Cornell Movie Dialogs corpus~\citep{DanescuNiculescuMizil2011ChameleonsII}, which consists of fictional conversations
from movie scripts.
We choose NLTK's Gutenberg dataset for our books portion, which consists of a subset of texts from Project Gutenberg~\citep{lebert2008project}.
Our news subset comes from the Gigaword dataset for abstractive summarization~\citep{graff2003english}. 
Lastly, our Wikipedia portion comes from WikiText-103~\citep{Merity2017PointerSM}.
For movies, news, and wiki, we extract sentences from its pre-specified validation set.
For books, since NLTK's Gutenberg dataset lacks a pre-specified data split, we consider the entire dataset.

\paragraph{Data Preprocessing}
We sentence tokenize all datasets using NLTK's sentence tokenizer.
To construct our dataset, we group sentences by sentence length into 8 bins: 5-10, 10-15, 15-20, 20-25, 25-30, 30-35, 35-40, and 40-128 using NLTK's word-level, regular expression tokenizer.
Next, we randomly sample 8 sentences from each bin to examine the efficacy of our method for a variety of sequence lengths.

\paragraph{Measuring the Effectiveness of Steering}
Given a target sentence $s$, we measure how well the steering vector $\zsteer$ can recover the target sentence by first greedily decoding from the language model with $\zsteer$, and then computing smoothed BLEU-4 using the target sentence $s$ and our decoded reconstruction $\hat{s}$~\cite{papineni2002bleu, chen2014systematic}.

\paragraph{Hyperparameter Search}
Our initial experiments showed little variation to most hyperparameters such as initialization method and learning rate schedule, so we fixed them in subsequent experiments using the values in Table~\ref{tbl:appendix_hyperparameters} in the appendix. 
We choose GPT2-117M as our language model and evaluate recovery on our dataset while varying injection locations and injection timesteps, the two hyperparameters that affect results significantly.
We present a subset of the results in Table~\ref{tbl:main_extraction} and the full set in the appendix (Tables~\ref{tbl: appendix_experiments},~\ref{tbl: appendix_experiments2}, and~\ref{tbl: appendix_experiments3}).

\begin{table}[t!]\small \centering
\begin{tabular}{@{}c|c|c@{}}
\toprule
\textbf{Injection location} & \textbf{Timestep} & \textbf{BLEU-4} \\ \midrule
Embedding & all timesteps & 33.99 \\ \midrule
\textbf{Layer 6 (self attn)} & \textbf{all timesteps} & \textbf{100.0} \\ \midrule
\textbf{Layer 6 (self attn)} & \textbf{first timestep} & \textbf{99.80} \\ \midrule
\textbf{Layer 7 (feed fwd)} & \textbf{all timesteps} & \textbf{100.0} \\ \midrule
\textbf{Layer 7 (feed fwd)} & \textbf{first timestep} & \textbf{99.25} \\ \midrule
\textbf{\makecell{All layers \\ (self attn + feed fwd)}} & \textbf{all timesteps} & \textbf{100.0} \\ \midrule
\textbf{\makecell{All layers \\ (self attn + feed fwd)}} & first timestep & 91.72 \\ \midrule
LM head & all timesteps & 6.72 \\ \bottomrule
\end{tabular}\caption{Sentence recovery for steering vectors when injected into different layers of the transformer model (Figure~\ref{fig:process}) and at multiple timesteps (all timesteps or first timestep). Results show that injecting a steering vector into the transformer stack, even at just the first timestep, can lead to nearly perfect recovery as long as it is in the middle of the network (layers 6 or 7 of 12).}  %
\label{tbl:main_extraction}
\end{table}
        
\subsection{Recovery effectiveness}
Table~\ref{tbl:main_extraction} shows reconstruction performance for several injection methods and indicates that we can recover a target sentence with perfect recovery when injecting $\zsteer$ in the middle of the transformer stack (layers 6 or 7 of 12) at just the first timestep and at all timesteps, for sequences up to 128 tokens.
We surmise that the middle layers of the transformer stack encode sufficiently rich feature representations that a small perturbation of a hidden layer, a steering vector, is sufficient to recover a sentence.
The success of steering vectors when injected in the middle of the transformer could help explain why adapter-based fine-tuning is effective.

In contrast, we find that we cannot steer GPT-2 at either the embedding or final language model head locations.
We suspect this is due to the fact that the embedding layer solely captures low-level information~\cite{lin-etal-2019-open, ethayarajh-2019-contextual, rogers-etal-2020-primer}.
Poor recovery at the LM head location is somewhat surprising, but could be explained by noting that the model has very low capacity above this layer. This suggests that alternative steering mechanisms, such as DExperts, that intervene at the output layers could potentially be improved by modifying hidden states elsewhere in the transformer stack~\cite{liu-etal-2021-dexperts}.

\begin{figure}[t]\centering
\includegraphics[width=0.42\textwidth]{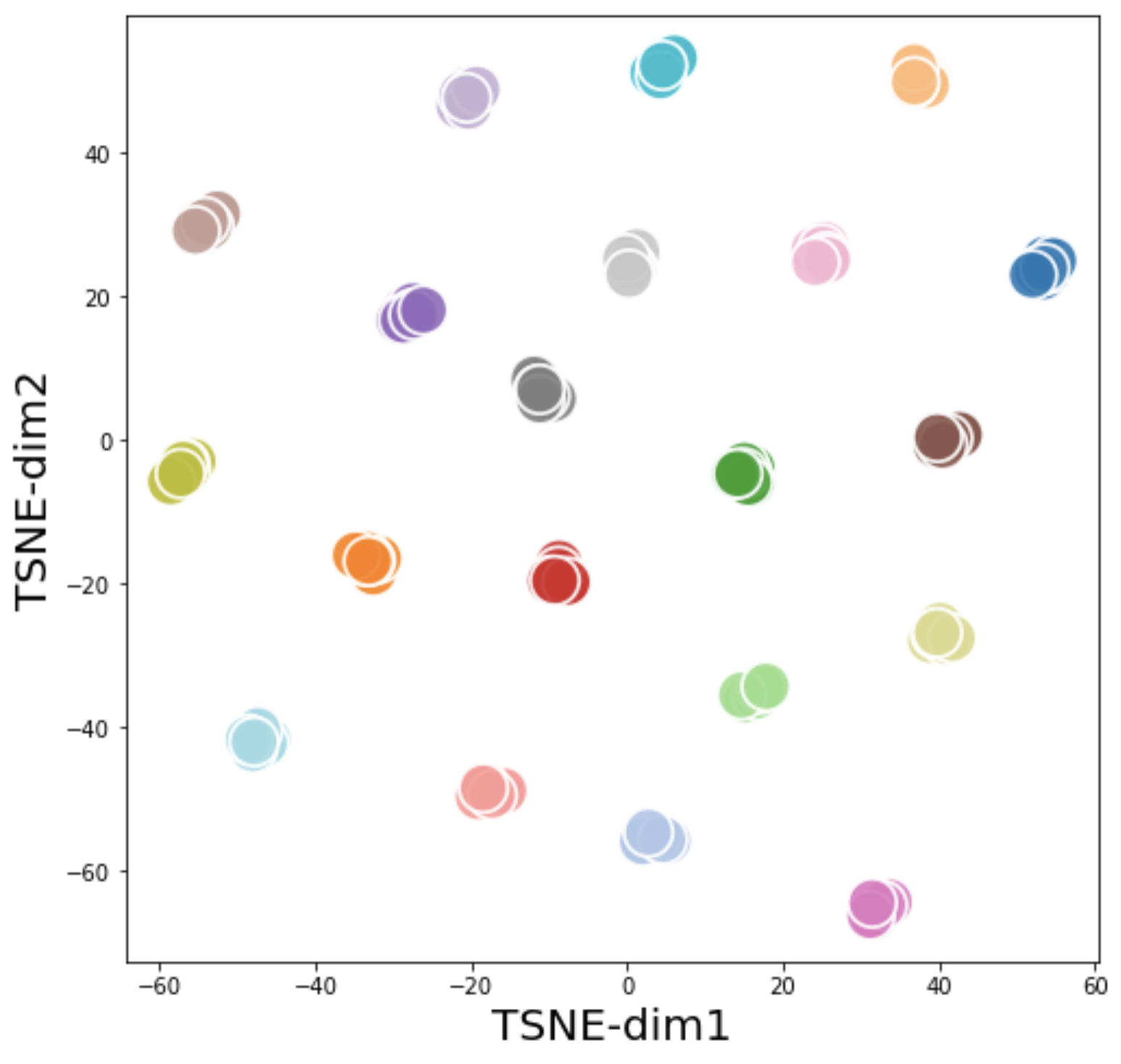}
\caption{TSNE projection of 8 steering vectors initialized from different random seeds for 20 different sentences (injected at layer 6, after self-attention).
$\zsteer$ is well-separated for different sentences, and the different seeds are tightly clustered for the same target sentence, indicating that our extraction method is robust.
}
\label{fig:distance_clusters}
\end{figure}

\paragraph{Robustness}
Now that we have established that steering vector extraction is possible, we explore whether there exist multiple steering vectors which recover the same sentence, and if so, what the relationship is between these vectors.
To do this, we take all 64 sentences from the books subset of the main dataset and initialize 8 different steering vectors for each sentence from different seeds.
Experiments reveal that for most sentences (63 of 64) all initializations recover the target sentence perfectly, confirming the robustness of our method. 

Latent geometry in text-based auto-encoders struggle with mapping vectors from one space to another consistently (e.g. token space to latent space)~\cite{bowman2016generating, Shen2020EducatingTA}.
The denoising auto-encoder offers a more consistent token space to latent space mapping~\cite{Vincent2008ExtractingAC}.
To explore whether our steering vectors have a distance-preserving mapping, we cluster the different initializations of steering vectors.
We extract 8 steering vectors for each of 20 sentences from the books corpus and down-project them into two-dimensions via TSNE~\cite{Maaten2008VisualizingDU}.
Figure~\ref{fig:distance_clusters} shows 20 distinct clusters, one for each sentence.
This indicates that distances between different vectors representing the same target sentence are much smaller than distances between vectors representing different sentences, and that distances in token space could be reflected in the latent steering space.

Motivated by the clustering results, we investigate whether the mean vector of the 8 extracted steering vectors for each target sentence recover the same sentence.
Experiments show that mean vectors are able to recover target sentences nearly perfectly, leading to a BLEU-4 of 99.4, further establishing the robustness of our method.

\section{Is unsupervised style transfer in the latent steering space possible?}
We explore whether vector arithmetic in this space is possible in the context of unsupervised style transfer.
In other words, we measure whether adding an offset vector, which captures the desired style transfer, to the steering vector effectively changes the style of the generated sentence.
Here, we show that unsupervised vector arithmetic with steering vectors is effective for unsupervised sentiment transfer, with performance comparable to models tailored to this task.

After extracting steering vectors for each sentence, we compute offset vectors by averaging steering vectors for a set of sentences in the source style $\Bar{z}_{source}$ and subtracting from the average of a set of steering vectors for the target style $\Bar{z}_{target}$.
Next, we flip the style of each sentence in our test set by adding the respective style transfer vector directly to its steering vector after scaling it by $\lambda$:
\begin{align}
    z_{totarget}&= \Bar{z}_{target} - \Bar{z}_{source}\\
    \hat{z}_{target} &= z_{source} + \lambda \cdot z_{totarget}
\end{align}

\begin{figure}[t!]\centering
\includegraphics[width=0.49\textwidth]{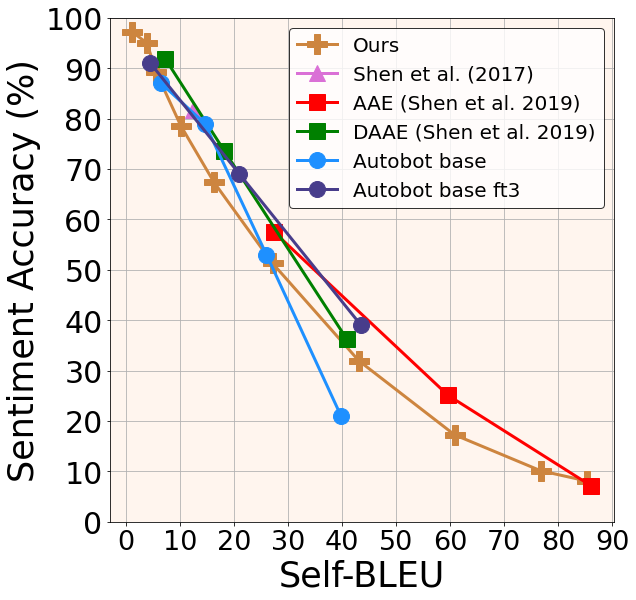}
\caption{Evaluation of unsupervised sentiment transfer on the Yelp dataset.
The plot shows accuracy vs. self-BLEU by varying $\lambda=$ (0.25, 0.5, 1.0, 1.5, 2.0, 2.5, 3.0, 4.0, 5.0, 10.0) for our method.
Overall, the steering vectors perform comparably to prior work.
}
\label{fig:sentiment_transfer}
\end{figure}

\paragraph{Unsupervised Sentiment Transfer}
Using the Yelp Sentiment dataset preprocessed by~\citet{Shen2017StyleTF}, we take 100 sentences from the validation set from each of the two classes of sentiment to compute offset vectors and  evaluate on the test set.
Following prior work~\cite{Shen2017StyleTF}, we measure how well this approach flips the sentiment of the sentence by measuring the accuracy of a RoBERTA-base model fine-tuned on the Yelp sentiment dataset.
We also measure the BLEU-4 between the style transferred sentences and the original and report the results in Figure~\ref{fig:sentiment_transfer}. We call this Self-BLEU following prior work.
For this experiment, our steering vectors are injected after the 7th self-attention layer at the first timestep.

\begin{table}[]
\small
\centering
\begin{tabular}{c|l}
\toprule
\multicolumn{2}{c}{\textbf{Steering vectors}} \\ \midrule
Positive Input & the taste is excellent!\\ 
+$0.5*z_{tonegative}$ & the taste is excellent!\\ 
+$1.0*z_{tonegative}$ & the taste is excellent!\\ 
+$1.5*z_{tonegative}$ & \makecell[l]{the taste is bitter and bitter\\taste is bitter taste is bitter} \\
+$2.0*z_{tonegative}$ & the taste is unpleasant.\\ \midrule
Negative Input & the desserts were very bland.\\ 
+$0.5*z_{topositive}$ & the desserts were very bland.\\ 
+$1.0*z_{topositive}$ & the desserts were very bland .\\ 
+$1.5*z_{topositive}$ & the desserts were very tasty.\\ 
+$2.0*z_{topositive}$ & the desserts were very tasty.\\ \bottomrule
\end{tabular}\caption{Examples of transferring sentiment using steering vectors for a positive input sentence (top) and negative input sentence (bottom). These results show fluency and accuracy in transfers while preserving the content of the input sentence.}\label{tbl: sent-transfer-examples}
\end{table}

We find that simple vector arithmetic via our steering vectors, which is fully unsupervised, performs comparably to~\citet{Shen2017StyleTF}, who learn an autoencoder-based model for the task in a fully supervised manner.
Our method also compares well with the Autobot~\cite{montero-etal-2021-sentence}, AAE, and DAAE models~\cite{Shen2020EducatingTA}, which although are unsupervised, either require training on in-domain data or require pretraining on millions of tokens in order to be effective. 
Other methods that use techniques from unsupervised machine translation to leverage the unpaired data in the task outperform all of these methods significantly~\cite{Hu2017TowardCG, Lample2019MultipleAttributeTR, He2020APF}. 
These methods are not directly comparable to ours, as they evaluate on a different test set altogether and use the training set to train directly.
Our method only requires access to 100 labeled examples per class to compute $\Bar{z}_{source}$ and $\Bar{z}_{target}$, far fewer than other baselines.
With as few as 10 examples per class, performance of our method remains competitive with autoencoder-based baselines.

Table~\ref{tbl: sent-transfer-examples} shows examples generated by our method for two input sentences.
We find that resulting sentences become more positive or negative with increasing $\lambda$ and often modify adjectives by swapping them out.
On closer inspection, we find that fluency is often challenging for higher values of $\lambda$ and that the generated sequences repeat individual words or phrases.
In addition, we find that negative to positive sentiment transfer is qualitatively more fluent and accurate than positive to negative sentiment transfer; see Table~\ref{tbl: appendix: sent-transfer-examples-extra} in the appendix for more example generations.
Lastly, we evaluate on 19 paired style transfer tasks from the StylePTB dataset~\cite{lyu-etal-2021-styleptb}, but modify the tasks to be unsupervised, following the same approach as above.
We find that our method is similarly effective on these tasks; see Table~\ref{tbl: appendix_styleptb} in the appendix for details.

\section{Do distances between steering vectors reflect sentence similarity?}
Previously, we found there exist multiple steering vectors that recover a target sentence and that those steering vectors are close together. 
This indicates the potential for distances in token space to be reflected in distances in the latent space occupied by steering vectors.
In this section, we explore whether distances relate to semantic similarity.
To do so, we use the STS-B test dataset, which consists of
sentence pairs and similarity scores. 
To evaluate our method we extract steering vectors for each sentence separately, compute cosine similarity,
and then correlate cosine similarity with annotator similarity via Spearman rank correlation.

\begin{figure}[t!]\centering
\includegraphics[width=0.49\textwidth]{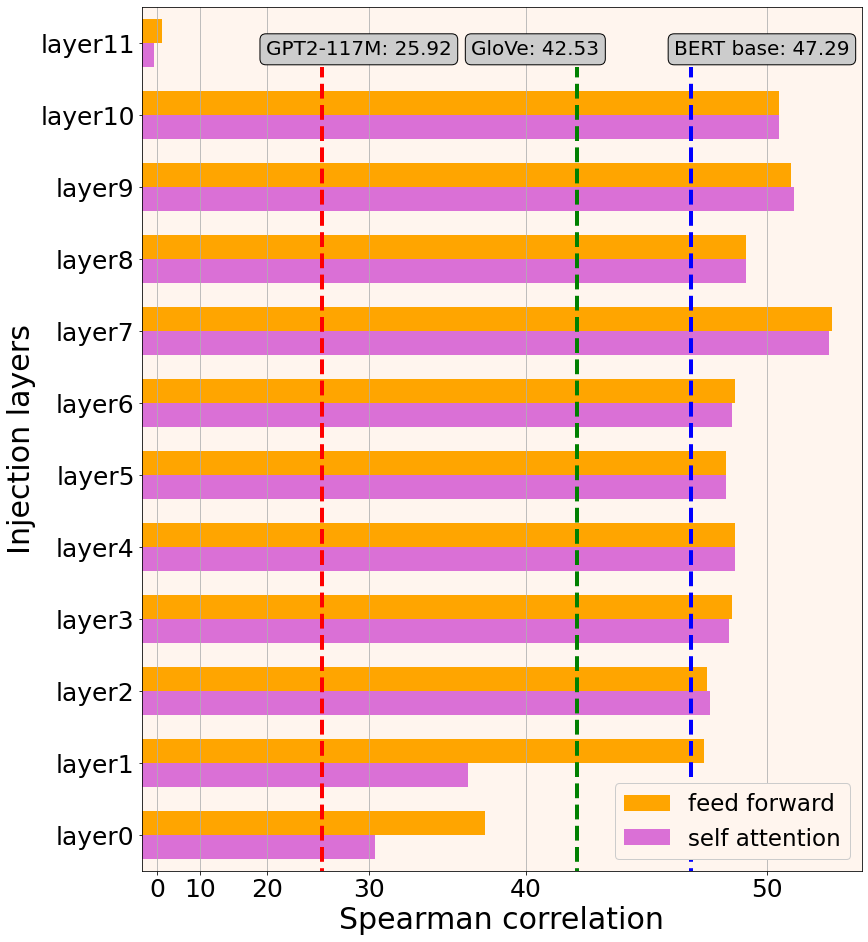}
\caption{On the test split of STS-B, we measure Spearman rank correlation ($\rho \cdot100$) between sentence similarity scores and cosine similarities between the steering vectors extracted from GPT2-117M when injected at different layers at the first timestep for those sentences. The vertical lines indicate extractive baselines: mean-pooled final hidden states for GPT2-117M and BERT-base as well as mean-pooled GloVe vectors. Results show that extracted steering vectors outperform these.}
\label{fig:stsb_layer_by_layer}
\end{figure}

In Figure~\ref{fig:stsb_layer_by_layer}, we show how well extracted steering vectors perform when injected at different layers and at the first timestep in the transformer stack.
This observation mirrors the results from the experiment on recovery effectiveness: middle layers in the transformer stack are ideal for steering, leading to perfect recovery and highest performance on semantic similarity. 
We outperform mean pooling the final hidden states of GPT2-117M and BERT-base as well as averaged GloVe vectors.
Even though our method is fully extractive, cosine distances reflect semantic similarity well.
We take our two best performing configurations, the 7th self-attention layer and the 7th feedforward layer, and compare with unsupervised methods for textual similarity.
Table~\ref{tbl: stsb} shows that our extracted steering vectors out-perform prior extractive unsupervised methods.
Predictably, however, methods which pretrain or fine-tune models on natural language inference datasets such as AutoBot~\cite{montero-etal-2021-sentence}, InferSent~\cite{conneau-etal-2017-supervised}, and SBERT~\cite{reimers-gurevych-2019-sentence} perform better. 
Lexical methods tailored for semantic similarity such as GloVe with uSIF-weighting and piecewise component removal (GloVE + UP; \citet{ethayarajh-2018-unsupervised}) and GloVe + WR~\cite{Arora2017ASB} also outperform our method.

\begin{table}[t!]
\small
\centering
\begin{tabular}{l|c|c}
\toprule
\textbf{Method} & \textbf{Spearman} & \textbf{Pearson} \\ \midrule
\multicolumn{3}{c}{\textit{Extractive methods}} \\ \midrule
Avg GPT2-117M embeddings & 25.92 & 16.52 \\ \midrule
Avg Bert embeddings & 47.29 & 47.91 \\ \midrule
Avg GloVe embeddings & 42.53 & 40.25 \\ \midrule
\textbf{Layer-7 self attention (ours)} & \textbf{52.04} & \textbf{51.17} \\ \midrule
\textbf{Layer-7 feedforward (ours)} & \textbf{52.08} & \textbf{51.18} \\ \midrule
\multicolumn{3}{c}{\textit{NLI-finetuned methods}} \\ \midrule
AutoBot-base & 58.49 & - \\ \midrule
InferSent - GloVe & 68.03 & - \\ \midrule
\textbf{SBERT-NLI-base} & \textbf{77.03} & - \\ \midrule
\multicolumn{3}{c}{\textit{Lexical methods}} \\ \midrule
GloVe+UP & - & 71.5 \\ \midrule
\textbf{GloVe+WR} & - & \textbf{72.0} \\ \bottomrule
\end{tabular}\caption{We evaluate performance on the STS-B test set by measuring Spearman rank correlation and Pearson correlation ($\rho \cdot100$). We take our two best performing configurations from Figure~\ref{fig:stsb_layer_by_layer} and compare them with three classes of unsupervised methods: extractive, NLI-finetuned, and lexical methods. Our method outperforms the extractive methods, but performs worse than the other methods, which are tailored for this task.}\label{tbl: stsb}
\end{table}

\section{Analysis of Properties}
\subsection{Interpolation}
\begin{figure}[t!]
\centering
\includegraphics[width=\columnwidth]{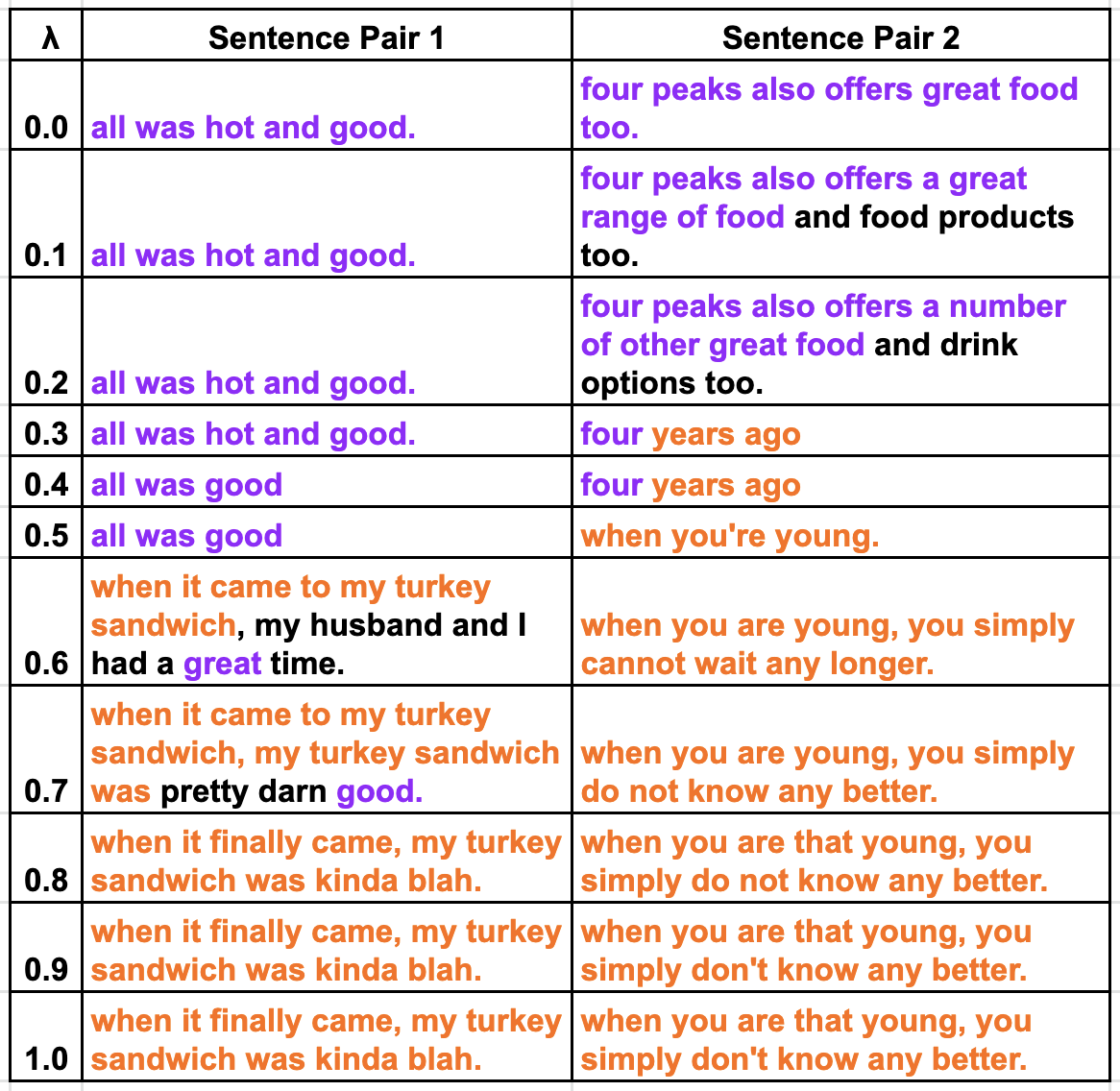}
\caption{Interpolation between steering vectors extracted from two pairs of random sentences from the Yelp Sentiment test set.  Decoding from interpolated vectors from two sentences produces well-formed output that incrementally changes the sentiment and meaning.
}
\label{fig:interpolation_both}
\end{figure}

Previous experiments indicate that the latent space occupied by steering vectors could be well-formed and smooth. 
To evaluate this qualitatively, we show linear interpolations of two pairs of steering vectors extracted from the Yelp Sentiment dataset in  Figure~\ref{fig:interpolation_both}.
The space between the vectors look smooth with well-formed grammatical sentences that mix the content of two sentences effectively.
The first interpolation (sentence pair 1) in Figure~\ref{fig:interpolation_both} shows that the positive sentiment of the first sentence carries all the way to $\lambda=0.7$, despite the content of the sentence changing to the second sentence.
The second interpolation (sentence pair 2) in Figure~\ref{fig:interpolation_both} indicates that the latent space could encode some semantics relating to time.
The second sentence includes the word "young" and so the transition between the two in $\lambda=0.3, 0.4$ combines the word "four" from the first sentence with the temporal component of "years ago" to relate the two sentences.
Lastly, for each individual sentence there exists a radius around it where those vectors also steer the language model to generate the same target sentence.
This could indicate that sentences have a representative volume from which, if any vector was sampled, could recover the sentence.

\subsection{Sampling}
Previous experiments show distances reflect semantic similarity and hint at the possibility that the latent space is smooth.
Given this, we evaluate whether we can sample from this space.
We take 4000 extracted steering vectors from the Yelp Sentiment test set.
We treat each dimension of the steering vector as an independent random variable that is normally distributed with a mean and variance equal to the mean and variance across that dimension over this set of steering vectors.
Table~\ref{tbl: appendix_sampling} shows the results of sampling 24 steering vectors and generating from them.
We observe mixed results: 5 samples lead to fully-formed sentences, and the remaining 19 lead to single tokens or phrases, indicating that treating steering vectors as samples from a multivariate Gaussian is not a reliable approach for sampling well-formed text.

\subsection{Intrinsic Dimension \& Space Complexity}
We define the intrinsic dimension of the task of steering a language model as the minimum dimension of $\zsteer$ that achieves perfect recovery on a set of sentences.
To measure intrinsic dimension, we vary the dimensions of $\zsteer$, choosing 192, 384, 576, 768.
We observe that reconstruction BLEU increases as the steering vector dimension increases, indicating that 768 dimensions may be needed to recover sequences nearly perfectly.
Given this, we conclude that the intrinsic dimension is at most 768.
However, a lower-dimensional representation can recover most sentences: 384 dimensions led to a reconstruction BLEU of 83.29.
See Table~\ref{tbl: intrinsic_dim} for more details.
Additionally, we find that sentence length and reconstruction BLEU are inversely correlated, i.e. longer sequences are harder to recover.
This is well-known; the number of bits needed to encode a sequence grows linearly with its length.
We find that all four dimensions of steering vectors can recover short sentences, but lower dimensional steering vectors struggle to recover longer ones.

\begin{table}[h!]
\small\centering
\begin{tabular}{l|l|l|l|l}
\toprule
\textbf{\makecell{Steering vector \\ dimension}} & 192 & 384 & 576 & 768 \\ \midrule
\textbf{\makecell{Reconstruction \\ BLEU-4}} & 43.43 & 83.29 & 93.93 & 100.00 \\ \bottomrule
\end{tabular}\caption{
Reconstruction BLEU for different steering vector dimensions.
Sentence recovery increases monotonically as the dimension increases, up to 100\% recovery at the model's hidden dimension.
}\label{tbl: intrinsic_dim}
\end{table}

Since steering vectors do not depend on sequence length, space complexity may not be a problem.
For a sequence of length 128, assuming 7 characters per word on average (including spaces), storage as a string takes $128*7=896$ bytes.
Our 768d steering vector uses 1536 bytes (fp16), but we can compress it by a factor of 2 (384d) sacrificing a little recovery (see Table~\ref{tbl: intrinsic_dim}) and store it using 768 bytes, less than its string representation.

\begin{figure}[t!]
\includegraphics[width=0.43\textwidth]{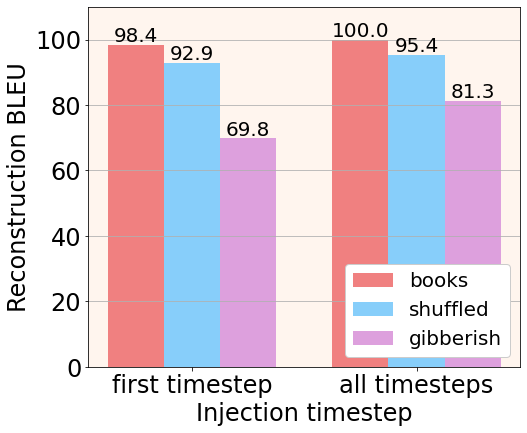}
\caption{We measure reconstruction BLEU for steering vectors learned for three datasets: books, shuffled, and gibberish.
Reconstruction BLEU for gibberish and shuffled data is lower than books indicating that the steering vector isn't just memorizing the sequence, but also leveraging the language model well.}
\label{fig:gibberish_shuffled}
\end{figure}

\subsection{Memorization}
Our nearly perfect recoverability performance indicates that steering vectors could either be encoding important properties by leveraging the language model, which would help generalization, or just simply be memorizing arbitrary sequences without using the underlying language model at all. 
In order to evaluate this, we randomly sample 64 sentences with lengths matching that of the books subset of our dataset, where each token is sampled uniformly at random with replacement from the vocabulary, and call this the gibberish fold of our dataset, following~\citet{subramani2019can}.
Secondly, to measure whether both content and word order affect recoverability, we construct another fold, the shuffled fold, by randomly shuffling the tokens in the sentences in the books subset.

Figure~\ref{fig:gibberish_shuffled} shows the results of injecting steering vectors into the 6th layer after the self-attention block in the transformer for all timesteps and the first timestep across all three datasets.
We observe that recoverability is highest for books, then shuffled, and lastly gibberish.
The gap between performance on books and gibberish indicates that steering vectors are not simply memorizing.
Since recovery on books is greater than recovery on shuffled, we conclude that steering vectors encode some information about word order.
Lastly, we notice that only passing the steering vector at the first timestep may reduce unwanted memorization capability because the relative difference in recovery between gibberish and the other sets is large.

\subsection{Connection to Prompting}
Motivated by the successes of prompt-based methods on zero-shot tasks with large generative language models such as GPT-3~\cite{Brown2020LanguageMA}, we evaluate a prompt-based version of our method.
Instead of adding $\zsteer$ to the hidden states of the language model, we concatenate $k$ steering vectors with the input embeddings, so that all tokens can attend to these $\zsteer$ vectors.
Experiments on the books subset show that recovery is much lower with this prompt-based approach than when injecting steering vectors directly into the transformer stack of the model. 
Even with $k=50$ steering vectors injected via this prompt-based approach, recovery fails to match that of a single steering vector $\zsteer$ injected into the hidden states of the language model.

\begin{table}[h!]
\small\centering
\begin{tabular}{c|c|c|c|c|c}
\toprule
\textbf{\makecell{Num prompt\\  vectors}} & 1 & 5 & 10 & 20 & 50 \\ \midrule
\textbf{\makecell{Reconstruction\\ BLEU-4}} & 81.7 & 94.3 & 98.7 & 98.6 & 98.5 \\ \bottomrule
\end{tabular}\caption{We measure reconstruction BLEU using a prompt-based approach, where latent steering vectors are concatenated to the embeddings. Even though each prompt vector is 768 dimensional, reconstruction BLEU is much lower in this setting than injecting a single steering vector into the layers of the transformer stack.}
\end{table}

\section{Related Work}
There exist many works, often using text-based autoencoders that try to induce a sentence representation space for controllable text generation by learning new models~\cite{Hu2017TowardCG, Shen2017StyleTF, Shen2020EducatingTA,
mai-etal-2020-plug,
montero-etal-2021-sentence}.
Our work concludes that we can extract steering vectors from pretrained models that have latent spaces that allow operations like this, without having to train any new models at all.
Other approaches control language models by adapting their hidden states using steerable layers, adapters, or steering their logits using auxiliary language models~\cite{gulcehre2015using, dathathri2019plug, Houlsby2019ParameterEfficientTL, Zhang2020SideTuningAB, liu-etal-2021-dexperts, krause-etal-2021-gedi-generative}.
Our method differs from all of these: we extract steering vectors directly from a language model and operate on the latent space occupied by these vectors, never fine-tuning any component of the model.
\citet{subramani2019can} investigate whether LSTM-based language models have sentence representations from which they can generate the original sentence. 
Although this premise relates to our first question: can we extract steering vectors, we extend far beyond that showing that vector arithmetic in the context of unsupervised style transfer is effective in our latent steering space.

\section{Conclusion}
In this paper we introduce a different approach to controllable text generation, where we extract latent steering vectors directly from a pretrained language model without fine-tuning.
Further, we find that our steering vectors lead to near perfect recovery on English sentences from a variety of domains.
We show that vector arithmetic can be used in the context of unsupervised style transfer on the Yelp sentiment dataset and StylePTB benchmark, performing comparably to models tailored to these tasks.
Experiments reveal that distances between steering vectors reflect sentence similarity when evaluated on STS-B, outperforming extractive methods.
Finally, we analyze properties of the steering vectors.
Our results indicate that we can control frozen pretrained language models effectively through their latent steering space. 

\section{Ethics Statement}
We introduce a new approach for controllable text generation by extracting vectors from a pretrained language model, leveraging information that is already encoded in the language model. 
Large pretrained models are known to be biased and our method of extracting steering vectors can reflect biases already present in these large pretrained language models~\cite{Bender2021OnTD}. 
The methods we present for controllable text generation could potentially be used for many downstream tasks such as unsupervised style transfer, abstractive summarization, and offensive content removal.
Unfortunately, this also means that this technology has the potential to be misused to perpetuate biases or generate offensive or toxic text. 

Our technology does not guarantee removal of toxic content, even in the case of unsupervised style transfer from toxic to nontoxic text.
To use this method, we encourage readers to first take steps to address biases that are already present in the underlying language model.
Further we recommend that this technology not be used in high-stakes settings, especially those where deployment of this technology could cause harm. %

\bibliography{anthology, custom}
\bibliographystyle{acl_natbib}

\newpage
~
\newpage
\appendix
\section{Appendix}
\label{sec:appendix}

\subsection{Extracting steering vectors}

In this section, we show the hyperparameter configurations used for extracting steering vectors from GPT2-117M. 
Table~\ref{tbl:appendix_hyperparameters} contains the list of final hyperparameters that we use to extract steering vectors for the different analyses in this paper. 
Table~\ref{tbl: appendix_experiments} shows the recovery performance of steering vectors when injected at different layers in the transformer stack on our compiled dataset.
These experiments reveal that injecting in the middle of the transformer stack either after the self attention layer or the feedforward layer leads to the highest BLEU-4 performance.
In fact, any layer other than the first or last layer achieves nearly perfect recovery.

In Table~\ref{tbl: appendix_experiments2} we look at recovery performance when injecting steering vectors at the embedding layer, transformer stack, and language modeling head, as well as different combinations of them. 
Injecting steering vectors at every layer in the transformer stack performed best. 
Table~\ref{tbl: appendix_experiments3} shows how recoverability changes with respect to how many timesteps $\zsteer$ is injected at.
Injecting at all timesteps performs negligibly better than injecting at just the first timestep. 

\begin{table}[h]\centering\small
\begin{tabular}{c|c}
\toprule
\textbf{Hyperparameters} & \textbf{Values} \\ \midrule
\textbf{Model} & GPT-2-117M\\ \midrule
\textbf{Max train steps} & 500 \\ \midrule
\textbf{\begin{tabular}[c]{@{}c@{}}Vector initialization\\ strategy\end{tabular}} & Xavier normal \\ \midrule
\textbf{Learning rate} & {[0.01, 1.0]} \\ \midrule
\textbf{Optimizer} & Adam \\ \midrule
\textbf{\begin{tabular}[c]{@{}c@{}}Learning rate\\ Scheduler\end{tabular}} & Decay on a plateau \\ \midrule
\textbf{\begin{tabular}[c]{@{}c@{}}Scheduler\\ decay factor\end{tabular}} & 0.9 \\ \midrule
\textbf{\begin{tabular}[c]{@{}c@{}}Scheduler\\ decay patience\end{tabular}} & 1.0 \\ \bottomrule
\end{tabular}\caption{List of hyperparameter configurations used to extract $\zsteer$ from GPT2-117M.}\label{tbl:appendix_hyperparameters}
\end{table}

\begin{table}[h]\centering\small
\begin{tabular}{l|c|c|c|c}
\toprule
\textbf{\makecell[l]{Injection\\location}} & \textbf{layers} & \textbf{timestep} & \textbf{lr} & \textbf{BLEU-4} \\ \midrule
self\_attn & 0 & all timesteps & 1 & 33.25 \\ \midrule
feedforward & 0 & all timesteps & 1 & 97.68 \\ \midrule
self\_attn & 1 & all timesteps & 1 & 98.06 \\ \midrule
feedforward & 1 & all timesteps & 1 & 99.54 \\ \midrule
self\_attn & 2 & all timesteps & 1 & 100.00 \\ \midrule
feedforward & 2 & all timesteps & 1 & 99.69 \\ \midrule
self\_attn & 3 & all timesteps & 1 & 100.00 \\ \midrule
feedforward & 3 & all timesteps & 1 & 100.00 \\ \midrule
self\_attn & 4 & all timesteps & 1 & 100.00 \\ \midrule
feedforward & 4 & all timesteps & 1 & 100.00 \\ \midrule
self\_attn & 5 & all timesteps & 1 & 100.00 \\ \midrule
feedforward & 5 & all timesteps & 1 & 100.00 \\ \midrule
self\_attn & 6 & all timesteps & 1 & \textbf{100.00} \\ \midrule
feedforward & 6 & all timesteps & 1 & 99.62 \\ \midrule
self\_attn & 7 & all timesteps & 1 & 99.62 \\ \midrule
feedforward & 7 & all timesteps & 1 & \textbf{100.00} \\ \midrule
self\_attn & 8 & all timesteps & 1 & 100.00 \\ \midrule
feedforward & 8 & all timesteps & 1 & 98.84 \\ \midrule
self\_attn & 9 & all timesteps & 1 & 99.22 \\ \midrule
feedforward & 9 & all timesteps & 1 & 98.61 \\ \midrule
self\_attn & 10 & all timesteps & 1 & 97.50 \\ \midrule
feedforward & 10 & all timesteps & 1 & 95.24 \\ \midrule
self\_attn & 11 & all timesteps & 1 & 86.04 \\ \midrule
feedforward & 11 & all timesteps & 1 & 6.29 \\ \bottomrule
\end{tabular}\caption{This table shows the reconstruction BLEU-4 for steering vectors from our compiled dataset when injected after different self attention and feedforward layers in the transformer stack. Injecting at the middle layer of the language model performs best.}\label{tbl: appendix_experiments}
\end{table}

\begin{table*}[h]\centering\small
\begin{tabular}{l|c|c|c}
\toprule
\textbf{Injection location} & \textbf{timestep} & \textbf{lr} & \textbf{BLEU-4} \\ \midrule
embedding & all timesteps & 0.01 & 33.99 \\ \midrule
every\_layer & all timesteps & 0.01 & 100.00 \\ \midrule
lm\_head & all timesteps & 0.01 & 6.72 \\ \midrule
embedding+every\_layer & all timesteps & 0.01 & 96.52 \\ \midrule
every\_layer+lm\_head & all timesteps & 0.01 & 100.00 \\ \midrule
embedding+lm\_head & all timesteps & 0.01 & 83.27 \\ \midrule
embedding+every\_layer+lm\_head & all timesteps & 0.01 & 98.11 \\ \midrule
every\_layer\_self\_attn & all timesteps & 0.01 & 99.62 \\ \midrule
\textbf{every\_layer+every\_layer\_self\_attn} & \textbf{all timesteps} & \textbf{0.01} & \textbf{100.00} \\ \midrule
every\_layer\_self\_attn+embedding+lm\_head & all timesteps & 0.01 & 97.31 \\ \midrule
every\_layer\_self\_attn+lm\_head & all timesteps & 0.01 & 99.62 \\ \midrule
every\_layer\_self\_attn+embedding & all timesteps & 0.01 & 94.28 \\ \bottomrule
\end{tabular}\caption{Here, we present the reconstruction BLEU-4 results for steering vectors on our multi-domain compiled dataset. We vary injection location here and observe that injecting into the transformer stack is necessary for good recovery. Injecting at the embedding or language model head performs poorly.}\label{tbl: appendix_experiments2}
\end{table*}

\begin{table*}[h]\centering\small
\begin{tabular}{l|c|c|c}
\toprule
\textbf{Injection location} & \textbf{timestep} & \textbf{lr} & \textbf{BLEU-4} \\ \midrule
every\_layer+every\_layer\_self\_attn & all timesteps & 0.01 & 100.0 \\ \midrule
every\_layer+every\_layer\_self\_attn & first timestep & 0.01 & 91.7 \\ \midrule
Layer 7 (feedforward) & all timesteps & 1 & 100.0 \\ \midrule
Layer 7 (feedforward) & first timestep & 1 & 99.2 \\ \midrule
Layer 6 (self\_attn) & all timesteps & 1 & 100.0 \\ \midrule
Layer 6 (self\_attn) & first timestep & 1 & 99.8 \\ \bottomrule
\end{tabular}\caption{In this table, we vary the timestep where we inject $\zsteer$ (all timesteps or first timestep) for three of our best injection locations. We again evaluate on our multi-domain compiled dataset and find that injecting at just the first timestep has a negligible decrease in recovery performance.}\label{tbl: appendix_experiments3}
\end{table*}

\begin{table*}[h]\small
\centering
\begin{tabular}{@{}c|c|c|c|c|c|c}
\toprule
\textbf{} & \textbf{Ours: $\lambda=0.25$} & \textbf{GPT2-finetune} & \textbf{Seq2seq} & \textbf{TAILOR} & \textbf{Neural QCFG + copy} & \textbf{Retrieve-Edit} \\ \midrule
\textbf{AAR} & \textbf{0.825} & - & - & - & - & - \\
\textbf{AEM} & \textbf{0.774} & 0.263 & 0.187 & - & 0.676 & 0.387 \\
\textbf{ARR} & 0.721 & 0.647 & 0.450 & 0.781 & - & \textbf{0.897} \\
\textbf{ASR} & \textbf{0.819} & - & - & - & - & - \\
\textbf{ATP} & 0.666 & 0.476 & 0.373 & 0.556 & \textbf{0.836} & 0.681 \\
\textbf{IAD} & \textbf{0.772} & 0.479 & 0.345 & - & - & 0.493 \\
\textbf{LFS} & \textbf{0.396} & - & - & - & - & - \\
\textbf{MFS} & \textbf{0.748} & - & - & - & - & - \\
\textbf{NAR} & \textbf{0.825} & - & - & - & - & - \\
\textbf{NSR} & \textbf{0.677} & - & - & - & - & - \\
\textbf{PFB} & 0.819 & 0.398 & 0.393 & \textbf{0.842} & - & 0.541 \\
\textbf{PPR} & 0.393 & 0.763 & 0.330 & 0.717 & - & \textbf{0.798} \\
\textbf{PTA} & 0.574 & 0.433 & 0.339 & - & - & \textbf{0.714} \\
\textbf{SBR} & 0.120 & 0.430 & 0.317 & - & - & \textbf{0.706} \\
\textbf{TFU} & 0.699 & 0.895 & 0.527 & 0.873 & - & \textbf{0.899} \\
\textbf{TPA} & 0.478 & 0.836 & 0.478 & 0.884 & - & \textbf{0.935} \\
\textbf{TPR} & 0.692 & 0.754 & 0.516 & 0.710 & - & \textbf{0.909} \\
\textbf{VEM} & 0.548 & 0.309 & 0.289 & - & \textbf{0.664} & 0.416 \\
\textbf{VSR} & \textbf{0.739} & - & - & - & - & - \\ \bottomrule
\end{tabular}\caption{In this table, we show performance on StylePTB. Although our method is unsupervised, we outperform GPT2-finetune and seq2seq on most tasks. For minimal edit tasks such as AEM, ARR, ATP, and PFB, we achieve comparable performance to TAILOR, Neural QCFG + copy, and Retrieve-Edit, which are models trained specifically for these types of tasks. Note: we obtain the numbers for GPT2-finetune, Seq2seq, and Retrieve-Edit from ~\cite{lyu-etal-2021-styleptb}, for TAILOR from~\cite{Ross2021TailorGA}, and for Neural QCFG+copy from~\cite{Kim2021SequencetoSequenceLW}.}\label{tbl: appendix_styleptb}
\end{table*}

\begin{table*}[h]\centering\small
\begin{tabular}{l|l}
\toprule
\multicolumn{2}{c}{\textbf{Sampled Sequences}} \\ \midrule
... & mobile\\ \midrule
wine.. & \makecell[l]{the first time that we've seen a team that looked\\good on paper.}\\ \midrule
peopled by. & \makecell[l]{Gathering around the world, we can all agree that\\the next step is to get our voices heard.}\\ \midrule
kitchen..... & x\\ \midrule
life & item link\\ \midrule
nomnomnomnom & appointments\\ \midrule
of & kitate.com\\ \midrule
\makecell[l]{We're going to make sure that we have a safe and\\secure environment for our employees.} & 3\\ \midrule
app & hotel\\ \midrule
racial & \makecell[l]{imagine a world where every day we see a new\\voice in our communities.}\\ \midrule
applify & \makecell[l]{(AAP) - The United States and its European allies\\are pressing ahead with plans to boost the number\\ of refugees arriving in the country from Iraq and\\Syria.\textbackslash{}n \textbackslash{}nThe United States and its allies are\\pressing ahead on the issue as they work to boost\\the number and scope of refugees arriving in Europe.} \\ \midrule
iv & the best.\\ \bottomrule
\end{tabular}\caption{Here we show results from our sampling experiment, where we treat steering vectors as samples from $d$ independent normally distributed random variables. We sample 24 steering vectors, pass them to GPT2-117M, and decode, resulting in the 24 generations presented here.}\label{tbl: appendix_sampling}
\end{table*}
\subsection{Unsupervised Sentiment Transfer}
\paragraph{Yelp Sentiment}
We also include generations from the unsupervised sentiment transfer experiment on the Yelp dataset. Table~\ref{tbl: appendix: sent-transfer-examples-extra} shows 8 more generations.
These generations highlight the same trends as before: with increasing $\lambda$, sentiment transfer strength increases.
We find that some generations do more than just flip the sentiment of the major adjective in the sentence such as adding the phrase "a great way to get a good laugh" in the 4th negative to positive generation when $\lambda=2.5$.

\paragraph{StylePTB}
For this study, we use 19 of 21 paired style transfer tasks from the StylePTB dataset~\cite{lyu-etal-2021-styleptb}, but modify the tasks to be unsupervised, following the same approach as sentiment transfer.
We randomly sample 100 sentences for each class from the training split for each of the style classes and use those to compute offset vectors. 
This offset vector is then added to the steering vector of the sentence to transfer style.
We follow the evaluation in~\citet{lyu-etal-2021-styleptb} because we have ground truth data and compare with fully supervised methods.
Experiments show that unsupervised vector arithmetic with steering vectors performs comparably using BLEU-1 to supervised methods designed for style transfer on tasks that require minimal edits (adjective emphasis (AEM), active to passive (ATP), information addition (IAD), and PP front to back (PFB)). We report BLEU-1 following prior work.
See Table~\ref{tbl: appendix_styleptb} for results on all 19 tasks.
Note \citet{lyu-etal-2021-styleptb} do not report any baseline numbers for AAR, ASR, LFS, MFS, NAR, NSR, and VSR for any of their models.

\begin{table*}[h]
\small
\centering
\begin{tabular}{c|l|c|l}
\toprule
\multicolumn{4}{c}{\textbf{Unsupervised sentiment transfer using steering vectors}} \\ \midrule
\multicolumn{2}{c}{\textbf{Positive to negative}} & \multicolumn{2}{c}{\textbf{Negative to positive}}\\ \midrule
input & i highly recommend this place! & input & my goodness it was so gross. \\ 
+$0.5*z_{tonegative}$ & i highly recommend this place! & +$0.5*z_{topositive}$ & my goodness it was so gross.\\ 
+$1.0*z_{tonegative}$ & i highly recommend this place! & +$1.0*z_{topositive}$ & my goodness it was so gross.\\ 
+$1.5*z_{tonegative}$ & i highly recommend this place! & +$1.5*z_{topositive}$ & my goodness it was so gross.\\
+$2.0*z_{tonegative}$ & i was very disappointed. & +$2.0*z_{topositive}$ & my goodness it was so good.\\ \midrule
input & \makecell[l]{it is always good to find quality\\local spots when traveling.} & input & \makecell[l]{went here for the first time to try\\something new ... bad idea.} \\ 
+$0.5*z_{tonegative}$ & \makecell[lt]{it is always good to find quality\\local spots when traveling.} & +$0.5*z_{topositive}$ & \makecell[lt]{went here for the first time to try\\something new.}
\\ 
+$1.0*z_{tonegative}$ & \makecell[lt]{it is always good to find quality\\local spots when traveling.} & +$1.0*z_{topositive}$ & \makecell[lt]{went here for the first time to try\\something new.}
\\ 
+$1.5*z_{tonegative}$ & \makecell[lt]{it is always good to find\\local spots when traveling.} & +$1.5*z_{topositive}$ & \makecell[lt]{went here for the first time to try\\something new.}\\
+$2.0*z_{tonegative}$ & it was always going to be a long time. & +$2.0*z_{topositive}$ & \makecell[lt]{went here for the first time to try\\something new. I'm really looking\\forward to trying something new\\for the first time.}\\ \midrule

input & it was delicious! & input & \makecell[lt]{if i could give them a zero\\star review i would!} \\ 
+$0.5*z_{tonegative}$ & it was delicious! & +$0.5*z_{topositive}$ & if i could give them a star i would!\\ 
+$1.0*z_{tonegative}$ & it was delicious! & +$1.0*z_{topositive}$ & if i could give them a star i would!\\ 
+$1.5*z_{tonegative}$ & it was a very bad night. & +$1.5*z_{topositive}$ & if i could give them a star i would! \\
+$2.0*z_{tonegative}$ & it was a very bad night. & +$2.0*z_{topositive}$ &  if i could give them a star i would!\\ \midrule

input & \makecell[lt]{the food is fresh and the\\environment is good.} & input & fries are n't worth coming back. \\ 
+$0.5*z_{tonegative}$ & \makecell[lt]{the food is fresh and the\\environment is good.} & +$0.5*z_{topositive}$ & fries are good.\\
+$1.0*z_{tonegative}$ & \makecell[lt]{the food is fresh and the\\environment is good.} & +$1.0*z_{topositive}$ & fries are good.\\
+$1.5*z_{tonegative}$ & \makecell[lt]{the food is fresh and the\\environment is good.} & +$1.5*z_{topositive}$ & fries are good.\\
+$2.0*z_{tonegative}$ & the food is bad. & +$2.0*z_{topositive}$ & fries are good.\\ 
+$2.5*z_{tonegative}$ & the food was produced in the past. & +$2.5*z_{topositive}$ & \makecell[lt]{fries are a great way to get a\\good laugh.}\\ \bottomrule
\end{tabular}\caption{This table shows some generations from unsupervised sentiment transfer of steering vectors. Sentences are from the Yelp dataset. We find that with increasing $\lambda$ sentiment transfers more strongly towards positive or negative, often switching at $\lambda=1.5$.}\label{tbl: appendix: sent-transfer-examples-extra}
\end{table*}

\subsection{Sampling}
In order to evaluate whether we can sample steering vectors reliably, we collect 4,000 extracted steering vectors from the Yelp Sentiment test set.
To generate, we consider each dimension of the steering vector as an independent random variable that is normally distributed.
The dimension means and variances are equal to the mean and variance for that dimension across this set of steering vectors.
In Table~\ref{tbl: appendix_sampling}, we show the results of sampling 24 steering vectors from these independent normally distributed random variables and generating from them using GPT2-117M as our language model. 
These results are mixed with approximately 20\% of the generations leading to fully formed sentences and the remaining 80\% corresponding to individual words or short phrases.
This could perhaps be partially explained by the fact that text from the web, including the corpora GPT2 was trained on, can often be of poor quality, especially when automatically crawled~\cite{Caswell2022QualityAA}.
Alternatively, our choice of considering $d$-dimensional steering vectors as samples from $d$ independent normally distributed random variables could be an incorrect assumption.
Alternative formulations could lead to more fluent and reliable generations.

\end{document}